\documentclass[conference]{IEEEtran}
\usepackage{amsthm}

\usepackage{amsmath}
\usepackage{amssymb}
\usepackage{amsfonts}
\usepackage{algorithm}
\usepackage{algorithmic}
\usepackage{array}
\usepackage{url}

\usepackage{commath}
\usepackage{color}
\usepackage{graphicx}
\usepackage{subfig}

\newtheorem{theorem}{Theorem}
\theoremstyle{plain}
\newtheorem{assumption}{Assumption}
\newtheorem{remark}{Remark}
\numberwithin{equation}{section}

\newcommand{\Real}{\mathbb{R}}
\newcommand{\Prox}[1]{\mbox{Prox}_{#1}}
\DeclareMathOperator*{\argmin}{arg\,min}

\ifCLASSINFOpdf
\else
\fi
\begin{document}

\title{Asynchronous Multi-Task Learning}


 \author{\IEEEauthorblockN{Inci M. Baytas\textsuperscript{1}, Ming Yan\textsuperscript{2,3}, Anil K. Jain\textsuperscript{1}, and Jiayu Zhou\textsuperscript{1}} 
\IEEEauthorblockA{\textsuperscript{1}Department of Computer Science and Engineering, Michigan State University, East Lansing, MI 48824\\
\textsuperscript{2}Department of Computational Mathematics, Science and Engineering, Michigan State University, East Lansing, MI 48824\\
\textsuperscript{3}Department of Mathematics, Michigan State University, East Lansing, MI 48824\\
Email: \{baytasin, myan, jain, jiayuz\}@msu.edu}
}

\maketitle

\begin{abstract}
Many real-world machine learning applications involve several
learning tasks which are inter-related. For example, in healthcare domain, we need to learn a 
predictive model of a certain disease for many hospitals. The models for each 
hospital may be different because of the inherent differences in the
distributions of the patient populations. However, the models are also closely related
because of the nature of the learning tasks modeling the same disease. By simultaneously learning all
the tasks, multi-task learning (MTL) paradigm performs inductive knowledge
transfer among tasks to improve the generalization performance. When datasets for the learning tasks are stored at different
locations, it may not always be feasible to transfer the data to
provide a data-centralized computing environment due to various practical
issues such as high data volume and privacy. In this paper, we propose a principled MTL framework
for distributed and asynchronous optimization to address the aforementioned challenges. In our framework, gradient
update does not wait for collecting the gradient information from all the tasks. Therefore, the proposed method is very efficient when the
communication delay is too high for some task nodes. We show that many
regularized MTL formulations can benefit from this framework, including the
low-rank MTL for shared subspace learning. Empirical studies on both synthetic and
real-world datasets demonstrate the efficiency and effectiveness of the
proposed framework.

\end{abstract}


\IEEEpeerreviewmaketitle

\section{Introduction}
\label{sect:intro}

Technological advancements in the last decade have transformed many industries, where the vast
amount of data from different sources of various types are collected and
stored. The huge amount of data has provided an unprecedented opportunity to
build data mining and machine learning algorithms to gain insights from the
data to help researchers study complex problems and develop better solutions.

In many application domains, we need to perform multiple machine learning tasks, where all the {\it tasks} involve learning a model
such as regression and classification. For example, in medical informatics, it
is of great interest to learn high-performance predictive models to accurately
predict diagnostic outcomes for different types of diseases from patients'
electronic medical records. Another example is to learn predictive
models for different hospitals, where hospitals are usually located in
different cities, and the distributions of patient populations are different from
hospital to hospital. Therefore, we will need to build one predictive model for
each hospital to address the heterogeneity in the population. In many
situations, such machine learning models are closely related. As in the
aforementioned examples, different types of diseases may be related through
similar pathways in the individual pathologies, and models from different
hospitals are inherently related because they are predicting the same target
for different populations. To leverage the relatedness among tasks, multi-task learning (MTL) techniques have been developed to simultaneously learn all the
related tasks to perform inductive knowledge transfer among the
tasks and improve the generalization performance. 

\begin{figure}[b!]
\center
\includegraphics[width=0.46\textwidth]{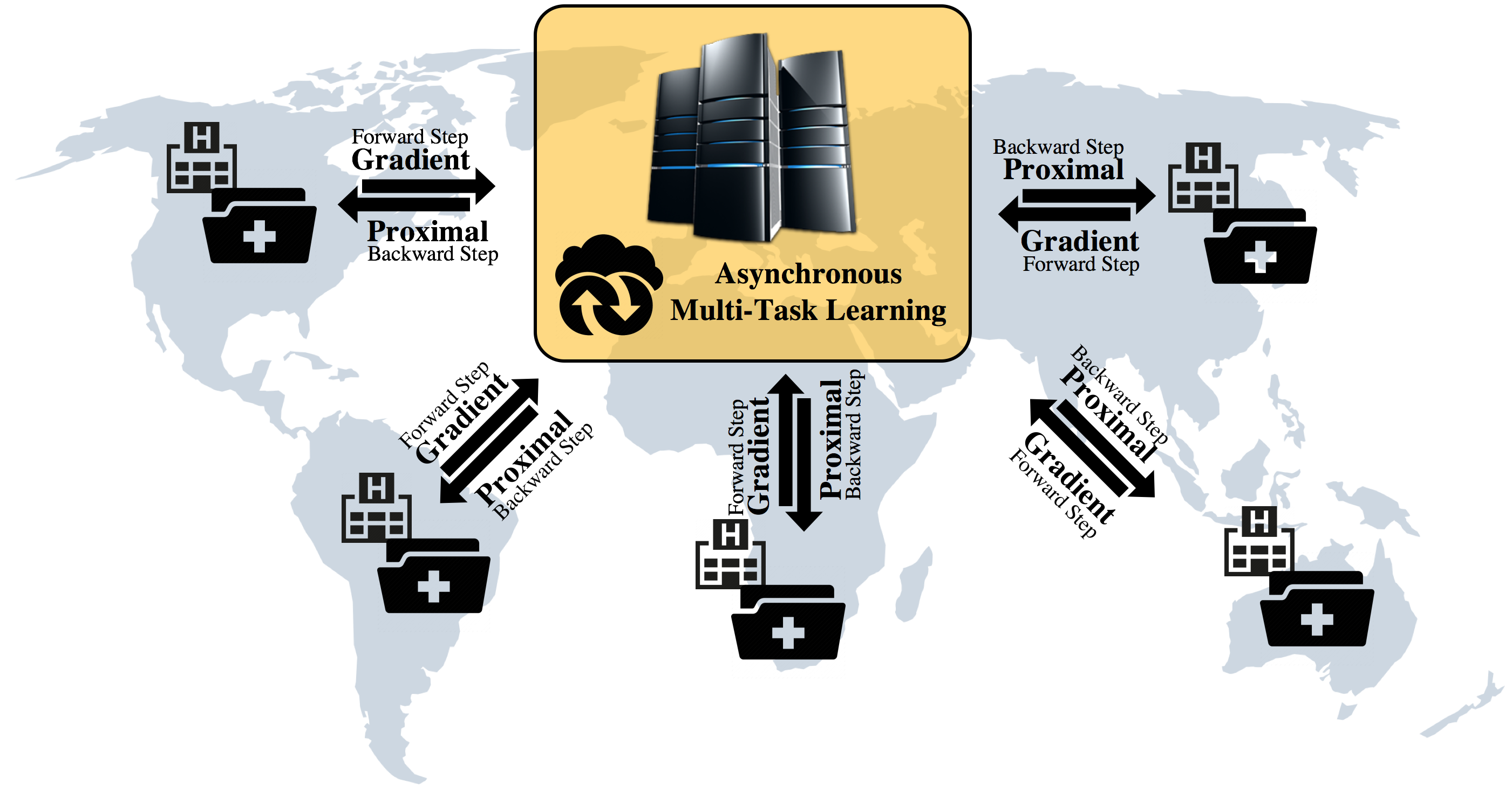}
\caption{Overview of the proposed asynchronous multi-task learning (AMTL) 
framework with applications to healthcare analytics. AMTL: i)
provides a principled way to solve the original MTL formulation in a
distributed manner; ii) has a linear convergence rate to find a global optimal solution for convex MTL
formulations under mild assumptions; iii) allows asynchronous updates from
multiple task nodes. }
\end{figure}

The rationale behind the MTL paradigm is that when there is not {\it enough
data} to learn a high-quality model, transferring and leveraging predictive
information from other related tasks can improve the generalization
performance. In the big data era, even though the {\it volume} of the data is
often huge, we may {\it still} not have enough data to learn high-quality
models because: (i) the number of features, hence the model parameters, can be
huge, which leads to the curse of dimensionality (e.g., tens of millions of
single nucleotide polymorphism in human genomic data); (ii) the distribution
of the data points in the training set can be highly skewed leading to bias in the 
learning algorithms (e.g., regional-biased patient population in each hospital); 
(iii) powerful models often have high model complexities or degree-of-freedom, and 
require much more training data for fitting (e.g., deep convolutional neural networks). 
Therefore, MTL is a critical machine learning technique for predictive 
modeling and data analytics even when high-volume data is available.

A majority of research on MTL has focused on regularized MTL, where
the task relatedness is enforced through adding a regularization term to the
multi-task loss function. The regularization term couples the learning tasks and induces
knowledge transfer during the learning phase. Regularized MTL is
considered to be powerful and versatile because of its ability to
incorporate various loss functions (e.g., least squares, logistic regression,
hinge loss), flexibility to model many common task relationships (e.g.,
shared feature subset, shared feature subspace, task clustering), and
principled optimization approaches. In order to couple the learning tasks, the
regularization term in regularized MTL is typically non-smooth, and thus renders 
the entire problem to be a composite optimization problem, which is typically 
solved by proximal gradient methods. 

Large-scale datasets are typically stored in distributed systems and
can be even located at different data centers. There are many practical
considerations that prevent the data from being transferred over the network. Let
us revisit the MTL in the setting of a hospital network, where the hospitals
are located in different regions and have their own data centers. The
patient data is extremely sensitive and cannot be frequently transferred over the
network even after careful encryption. As such, distributed algorithms are desired to
solve the distributed MTL problems, where each task node (server) computes the summary
information (e.g., gradient) locally and then the information from different nodes
is aggregated at a central node to guide the gradient descent direction. However,
because of the non-smoothness nature of the objective, the proximal projection in the MTL
optimization requires synchronized gradient information from all task nodes before
the next search point can be obtained. Such synchronized distributed
optimization can be extremely slow due to high network communication delays
in some task nodes.

In this paper, we propose a distributed MTL framework to
address the aforementioned challenges. The proposed MTL framework: 1)
provides a principled way to solve the original MTL formulation, as compared
to some recent work on distributed MTL systems that use heuristic or
approximate algorithms; 2) is equipped with a linear convergence rate for
a global optimal for convex MTL formulations under mild assumptions; 3) allows
asynchronized updates from multiple servers and is thus robust against network
infrastructures with high communication delays between the central node and some task nodes. The framework is capable of solving most existing
regularized MTL formulations and is compatible with the formulations in the
MTL package MALSAR~\cite{zhou2011malsar}. As a case study, we elaborate the low-rank MTL
formulations, which transfer knowledge via learning a low-dimensional subspace from task models. We evaluate the proposed framework on both
synthetic and real-world datasets and demonstrate that the proposed algorithm
can significantly outperform the synchronized distributed MTL, under different
network settings.

The rest of the paper is organized as follows: Section~\ref{sect:related-work}
revisits the major areas that are related to this research and establish the
connections to the state-of-the-arts. Section~\ref{sect:method} elaborates the
proposed framework. Section~\ref{sect:exp} presents the results from empirical
studies. Section~\ref{sect:conclusion} provides discussions on some key issues
and lays out directions for future work.

\section{Related Work}
\label{sect:related-work}

In this section, a literature review of distributed optimization and distributed MTL is given. 

\subsection{Distributed optimization}

Distributed optimization techniques exploit technological improvements in hardware to solve massive optimization problems fast. One commonly used distributed optimization approach is alternating direction method of multipliers (ADMM), which was firstly proposed in the 1970s~\cite{boyd2011}. Boyd {\it et al.} defined it as a well suited distributed convex optimization method. In the distributed ADMM framework in~\cite{boyd2011}, local copies are introduced for local subproblems, and the communication between the work nodes and the center node is for the purpose of consensus. Though it can fit in a large-scale distributed optimization setting, the introduction of local copies increases the number of iterations to achieve the same accuracy. Furthermore, the approaches in~\cite{boyd2011} are all synchronized.

In order to avoid introducing multiple local copies and introduce asynchrony, Iutzeler {\it et al.} proposed an asynchronous distributed approach using randomized ADMM~\cite{Franck2013} based on randomized Gauss-Seidel iterations of a Douglas-Rachford splitting (DRS) because ADMM is equivalent to DRS. For the asynchronous distributed setting, they assumed a set of network agents, where each agent has an individual cost function. The goal was to find a consensus on the minimizer of the overall cost function which consists of individual cost functions. 

Aybat {\it et al.} introduced an asynchronous distributed proximal gradient method by using the randomized block coordinate descent method in~\cite{Aybat2014} for minimizing the sum of a smooth and a non-smooth functions. The proposed approach was based on synchronous distributed first-order augmented Lagrangian (DFAL) algorithm. 


Liu {\it et al.} proposed an asynchronous stochastic proximal coordinate descent method in~\cite{Liu2015}. The proposed approach introduced a distributed stochastic optimization scheme for composite objective functions. They adopted an inconsistent read mechanism where the elements of the optimization variable may be updated by multiple cores during being read by another core. Therefore, cores read a hybrid version of the variable which had never been existed in the memory. It was shown that the proposed algorithm has a linear convergence rate for a suitable step size under the assumption that the optimal strong convexity holds.     

Recently, Peng {\it et al.} proposed a general asynchronous parallel framework for coordinate updates based on solving fixed-point problems with non-expansive operators~\cite{Arock,TMAC}. Since many famous optimization algorithms such as gradient descent, proximal gradient method, ADMM/DRS, and primal-dual method can be expressed as non-expansive operators. This framework can be applied to many optimization problems and monotone inclusion problems~\cite{Coordinate_update}. The procedure for applying this framework includes transferring the problem into a fixed-point problem with a non-expansive operator, applying ARock on this non-expansive operator, and transferring it back to the original algorithm. Depending on the structures of the problem, there may be multiple choices of non-expansive operators and multiple choices of asynchronous algorithms. These asynchronous algorithms may have very different performance on different platforms and different datasets.  

\subsection{Distributed multi-task learning}

In this section, studies about distributed MTL in the literature are summarized. In real-world MTL problems, geographically distributed vast amount of data such as healthcare datasets is used. Each hospital has its own data consisting of information such as patient records including diagnostics, medication, test results, etc. In this scenario, two of the main challenges are network communication and the privacy. In traditional MTL approaches, data transfer from different sources is required. However, this task can be quite costly because of the bandwidth limitations. Another point is that hospitals may not want to share their data with others in terms of the privacy of their patients. Distributed MTL provides a solution by using distributed optimization techniques for the aforementioned challenges. In distributed MTL, data is not needed to be transferred to a central node. Since only the learned models are transferred instead of the whole raw data, the cost of network communication is reduced. In addition, distributed MTL mitigates the privacy problem, since each worker updates their models independently. For instance, Dinuzzo and Pillonetto in~\cite{client11} proposed a client-server MTL from distributed datasets in 2011. They designed an architecture that was simultaneously solving multiple learning tasks. In their setting, client was an individual learning task. Server was responsible for collecting the data from clients to encode the information in a common database in real time. In this setting, each client could access the information content of all the data on the server without knowing the actual data. MTL problem was solved by regularized kernel methods in that paper.     

Mateos-N$\acute{\mbox{u}}\tilde{\mbox{n}}$ez and Cort$\acute{\mbox{e}}$z focused on distributed optimization techniques for MTL in~\cite{nunez15}. Authors defined a separable convex optimization problem on local decision variables where the objective function comprises a separable convex cost function and a joint regularization for low rank solutions. Local gradient calculations were divided into a network of agents. Their second solution consisted of a separable saddle-point reformulation through Fenchel conjugation of quadratic forms. A separable min-max problem was derived, and it has iterative distributed approaches that prevent from calculating the inverses of local matrices. 

Jin {\it et al.}, on the other hand, proposed collaborating between local and global learning for distributed online multiple tasks in~\cite{online15}. Their proposed method learned individual models with continuously arriving data. They combined MTL along with distributed learning and online learning. Their proposed scheme performed local and global learning alternately. In the first step, online learning was performed locally by each client. Then, global learning was done by the server side. In their framework, clients still send a portion of the raw data to the global server and the global server coordinates the collaborative learning.


In 2016, Wang {\it et al.} also proposed a distributed scheme for MTL with shared representation. They defined shared-subspace MTL in a distributed multi-task setting by proposing two subspace pursuit approaches. In their proposed setting, there were $m$ separate machines and each machine was responsible for one task. Therefore, each machine had access only for the data of the corresponding task. Central node was broadcasting the updated models back to the machines. As in~\cite{nunez15}, a strategy for nuclear norm regularization that avoids heavy communication was investigated. In addition, a greedy approach was proposed for the subspace pursuit which was communication efficient. Optimization was done by a synchronous fashion. Therefore, workers and master nodes had to wait for each other before proceeding.  

The prominent result is that all these methods follow a synchronize approach while updating the models. When there is a data imbalance, the computation in the workers which have larger amount of data will take longer. Therefore, other workers will have to wait, although they complete their computations. In this paper, we propose to use an asynchronous optimization approach to prevent workers from waiting for others and to reduce the training time for MTL.

\section{Asynchronous Multi-Task Learning}
\label{sect:method}

\subsection{Regularized multi-task learning}
In MTL, multiple related learning tasks are involved, and the goal of
MTL is to obtain models with improved generalization performance for {\it all} the tasks involved. Assume that we have $T$ supervised learning tasks, and for each
task, we are given the training data $\mathcal D_t = \{x_t, y_t\}$ of $n_t$ data
points, where $x_t \in \Real^{n_t \times d}$ is the feature vectors for the
training data points and $y_t \in \Real^{n_t}$ includes the corresponding
responses. Assume that the target model is a linear model parametrized by the vector
$w_t \in \Real^d$ (in this paper we slightly abuse the term ``model'' to denote
the vector $w_t$), and we use $\ell_t(w_t)$ to denote the loss
function $\ell_t(x_t, y_t; w_t)$, examples of which include the least squares loss and the logistic loss. In addition, we assume that the tasks can be heterogeneous~\cite{yang2009heterogeneous},
i.e., some tasks can be regression while the other tasks are classification. In a
single learning task, we treat the task independently and minimize the
corresponding loss function, while in MTL, the tasks are related, and we hope
that by properly assuming task relatedness, the learning of one task (the
inference process of $w_t$) can benefit from other tasks~\cite{Caruana97}. Let $W = [w_1, \dots,
w_T] \in \Real^{d \times T}$ collectively denote the learning parameters from
all $T$ tasks. We note that simply minimizing the joint objective
$f(W) =\sum\limits_{t=1}^T \ell_t(w_t)$ cannot achieve the desired
knowledge transfer among tasks because the minimization problems are
decoupled for each $w_t$. Therefore, MTL is typically achieved by adding 
a penalty term~\cite{Theo2004,zhou2011malsar,zhou2011clustered}:
\begin{align}
\min_{W} \left\{ \sum\nolimits_{t=1}^T \ell_t(w_t) + \lambda g(W) \right\} 
 \equiv f(W) + \lambda g(W), \label{eqt:mtl}
\end{align}
where $g(W)$ encodes the assumption of task relatedness and couples $T$ tasks, 
and $\lambda$ is the regularization parameter controlling how much knowledge 
to shared among tasks. In this paper, we assume that the loss function $f(W)$ is convex and $L$-Lipschitz differentiable with $L>0$ and $g(W)$ is closed proper convex. 
One representative task relatedness is joint feature learning, which is
achieved via the grouped sparsity induced by penalizing the $\ell_{2,1}$-norm~\cite{JunLiu09}
of the model matrix $W$: $g(W) = \|W\|_{2,1} = \sum_{i=1}^d \|w^i\|_2$ where
$w^i$ is the $i$-th row of $W$. The grouped sparsity would encourage many
rows of $W$ to be zero and thus remove the effects of the corresponding features on the
predictions in linear models. Another commonly used MTL method is the shared
subspace learning~\cite{argyriou2008convex}, which is achieved by penalizing
the nuclear norm $g(W) = \|W\|_* = \sum_{i=1}^{\min(d, T)} \sigma_i(W)$, where
$\sigma_i(W)$ is the $i$-th singular value of the matrix $W$. Intuitively, a
low-rank $W$ indicates that the columns of $W$ (the models of tasks) are
linearly dependent and come from a shared low-dimensional subspace. The nuclear
norm is the tightest convex relaxation of the rank
function~\cite{fazel2001rank}, and the problem can be solved via proximal gradient
methods~\cite{ji2009accelerated}.

\vspace{-0.05in}
\subsection{(Synchronized) distributed optimization of MTL}
Because of the non-smooth regularization $g(W)$, the composite objectives in
MTL are typically solved via the proximal gradient based first order
optimization methods such as FISTA~\cite{beck2009fast},
SpaRSA~\cite{wright2009sparse}, and more recently, second order proximal
methods such as PNOPT~\cite{lee2014proximal}. Below we review the two key
computations involved in these algorithms:

\vspace{+0.03in}
\noindent {\it 1) Gradient Computation}. The gradient of the smooth component 
is computed by aggregating gradient vectors from the loss function of each task: 
\begin{align}
\nabla & f(W) =
\nabla \sum\nolimits_{t=1}^T \ell_t(w_t) = [\nabla\ell_1(w_1), \dots, \nabla\ell_T(w_T) ]. \label{eqt:pg_gradient}
\end{align}

\vspace{+0.03in}
\noindent {\it 2) Proximal Mapping}. After the gradient update, the next search 
point is obtained by the proximal mapping which solves the following optimization problem:
\begin{align}
\Prox{\eta\lambda g}(\hat W) = \argmin_W \frac{1}{2\eta}\|W - \hat W\|_F^2 + \lambda g(W),
\label{eqt:pg_proximal}
\end{align}
where $\eta$ is the step size, $\hat W$ is computed from the gradient descent with step size $\eta$:
$\hat W = W - \eta \nabla f(W)$, and $\|\cdot\|_F$ is the Frobenius norm.

When the size of data is big, the data is typically stored in different pieces
that spread across multiple computers or even multiple data centers. For MTL,
the learning tasks typically involve different sets of samples, i.e., $\mathcal
D_1, \dots, \mathcal D_T$ have different data points, and it is very common
that these datasets are collected through different procedures and stored at
different locations. It is not always feasible to transfer the relevant data pieces
to one center to provide a centralized data environment for optimization. The reason why a centralized data environment is difficult to achieve is simply
because the size of the dataset is way too large for efficient network
transfer, and there would be concerns such as network security and data privacy.
Recall the scenario in the introduction, where the learning involves patients'
medical records from multiple hospitals. Transferring patients' data outside
the respective hospital data center would be a huge concern, even if the data is
properly encrypted. Therefore, it is imperative to seek distributed
optimization for MTL, where summarizing information from data is computed only
locally and then transferred over the network. 

Without loss of generality, we assume a general setting where the datasets
involved in tasks are stored in different computer systems that are connected
through a star-shaped network. Each of these computer systems, called a {\it node}, {\it
worker} or an {\it agent}, has full access to the data $\mathcal D_t$ for one
task, and is capable of numerical computations (e.g., computing gradient). We
assume that there is a {\it central server} that collects the information
from the task agents and performs the proximal mapping. We now investigate
aforementioned key operations involved and see how the optimization can be
distributed across agents to fit this architecture and minimize the
computational cost. Apparently, the gradient computation in
Eq.~(\ref{eqt:pg_gradient}) can be easily parallelized and distributed because
of the independence of gradients among tasks. Naturally, the proximal mapping
in Eq.~(\ref{eqt:pg_proximal}) can be carried out as soon as the gradient
vectors are collected and $\hat W$ is obtained. The projected solution after
the proximal mapping is then sent back to the task agents to prepare for the next
iteration. This {\it synchronized} distributed approach assembles a map-reduce
procedure and can be easily implemented. The term ``synchronize'' indicates that,
at each iteration, we have to wait for all the gradients to be collected before the
server (and other task agents) can proceed.

Since the server waits for {\it all} task agents to finish at every iteration,
one apparent disadvantage is that when one or more task agents are suffering from
high network delay or even failure, all other agents must wait. Because the
first-order optimization algorithm typically requires many iterations to
converge to an acceptable precision, the extended period of waiting time in a
synchronized optimization will lead to prohibitive algorithm running
time and a waste of computing resources. 


\subsection{Asynchronized framework for distributed MTL}

To address the aforementioned challenges in distributed MTL, we propose to
perform asynchronous multi-task learning (AMTL), where the central server begins to
update model matrix $W$ after it receives a gradient computation from one
task node, without waiting for the other task nodes to finish their computations.
While the server and all task agents maintain their own copies of $W$ in the memory,
the copy at one task node may be different from the copies at other nodes. The
convergence analysis of the proposed AMTL framework is backed up by a recent
approach for asynchronous parallel coordinate update problems by using
Krasnosel'skii-Mann (KM) iteration~\cite{Arock,TMAC}. 
A task node is said to be {\it activated} when it performs computation and network
communication with the central server for updates. The framework is based on 
the following assumption on the activation rate: 

\begin{assumption}\label{asp:act_rate}
All the task nodes follow independent Poisson processes and
have the same activation rate.
\end{assumption}

\begin{remark}\label{rmk:act_rate}
We note that when the activation rates are different for different task nodes,
we can modify the theoretical result by changing the step size: if a task
node's activation rate is large, then the probability that this task node is activated is large and thus the corresponding step size should be small. In
Section~\ref{sect:method:step_size} we propose a dynamic step size strategy
for the proposed AMTL.
\end{remark}


\begin{algorithm*}[t!]
\small
\caption{The proposed Asynchronous Multi-Task Learning framework}
\label{alg:amtl}
\begin{algorithmic}
\REQUIRE Multiple related learning tasks reside at task nodes, including the training data and the loss
function for each task $\{x_1, y_1, \ell_t\},...,\{x_T, y_T, \ell_T\}$, maximum delay $\tau$, step size $\eta$, multi-task regularization parameter $\lambda$.
\ENSURE Predictive models of each task $v_{1},...,v_{T}$.
\STATE Initialize task nodes and the central server.
\STATE Choose $\eta_k \in [\eta_{\min}, \frac{c}{2\tau/\sqrt{T}+1} ]$ for any
	constant $\eta_{\min}>0$ and $0<c<1$
\WHILE{ \textit{every task node asynchronously and continuously} } 
	\STATE Task node $t$ requests the server for the forward step computation $\mbox{Prox}_{\eta \lambda g}\left(\hat{{v}}^{k}\right)$, and 
	\STATE Retrieves 
	 $\left(\mbox{Prox}_{\eta \lambda g}\left(\hat{{v}}^{k}\right)\right)_t$ from the central server and 
	\STATE Computes the coordinate update on $v_t$
	\begin{equation}
	{v}_t^{k+1} = v_t^k + \eta_k \left(\left(\mbox{Prox}_{\eta \lambda g}\left(\hat{{v}}^{k}\right)\right)_t-\eta  \nabla \ell_t \left(\left(\mbox{Prox}_{\eta \lambda g} (\hat{{v}}^{k})\right)_t\right)-v_t^k\right)
	\label{eq:update}
	\end{equation}
	\STATE Sends updated ${v}_t$ to the central node. 
\ENDWHILE	
\end{algorithmic}
\end{algorithm*}

The proposed AMTL uses a backward-forward operator splitting method
\cite{combettes2005signal,Coordinate_update} to solve problem~\eqref{eqt:mtl}. Solving
problem~\eqref{eqt:mtl} is equivalent to finding the optimal solution $W^*$
such that $0\in \nabla f(W^*)+ \lambda \partial g(W^*)$, where $\partial g(W^*)$ denotes the set of subgradients of non-smooth 
function $g(\cdot)$ at $W^*$ and we have the
following:
\begin{align*}
 0\in  \nabla f(W^*) & + \lambda \partial g(W^*)
    \iff   -\nabla f(W^*)\in \lambda \partial g(W^*)  \\
    \iff & W^* -\eta \nabla f(W^*)\in W^* +\eta \lambda \partial g(W^*).
\end{align*}
Therefore the forward-backward iteration is given by:
\begin{align*}
W^+=(I+\eta \lambda\partial  g)^{-1}(I-\eta \nabla f)(W),
\end{align*}
which converges to the solution if $\eta\in (0,2/L)$. Since $\nabla
f(W)$ is separable, i.e., $\nabla f(W) = [\nabla \ell_1(w_1), \nabla
\ell_2(w_2),\cdots,\nabla \ell_T(w_T)]$, the forward operator, i.e.,
$I-\eta\nabla f$, is also separable. However, the backward operator, i.e.,
$(I+\eta\lambda\partial g)^{-1}$, is not separable. Thus, we can not apply
the coordinate update directly on the forward-backward iteration. If we switch the
order of forward and backward steps, we obtain the following backward-forward iteration:
\begin{align*}
V^+=(I-\eta \nabla f)(I+\eta \lambda \partial g)^{-1}(V),
\end{align*}
where we use an auxiliary matrix $V\in\Real^{d\times T}$ instead of $W$ during
the update. This is because the update variables in the forward-backward and
backward-forward are different variables. Moreover, one additional backward step is
needed to obtain $W^*$ from $V^*$. We can thus follow~\cite{Arock} to perform
task block coordinate update at the backward-forward iteration, where each
{\it task block} is defined by the variables associated to a task. The update
procedure is given as follows:
\begin{align*}
v_t^+=(I-\eta \nabla \ell_t)\left((I+\eta \lambda \partial g)^{-1}(V)\right)_t,
\end{align*}
where $v_t \in \Real^d$ is the corresponding auxiliary variable of $w_t$ for
task $t$. Note that updating one task block $v_t$ will need one full backward
step and a forward step only on the task block. The overall AMTL algorithm 
is given in~\ref{alg:amtl}. The formulation in Eq.~\ref{eq:update} is the update rule of 
KM iteration discussed in~\cite{Arock}. KM iteration provides a generic framework to solve fixed point 
problems where the goal is to find the fixed point of a nonexpansive operator. In the problem 
setting of this study, backward-forward operator is our fixed-point operator.
We refer to Section 2.4 of~\cite{Arock} to see how Eq.~\ref{eq:update} is derived.

In general, the choice between forward-backward or backward-forward is largely
dependent on the difficulty of the sub problem. If the backward step is
easier to compute compared to the forward step, e.g., data $(x_t,y_t)$ is large,
then we can apply coordinate update on the backward-forward iteration.
Specifically in the MTL settings, the backward step is given by a proximal
mapping in Eq.~\ref{eqt:pg_proximal} and usually admits an analytical solution
(e.g., soft-thresholding on singular values for trace norm). On the other hand, the gradient computation in
Eq.~\ref{eqt:pg_gradient} is typically the most time consuming step for large datasets. Therefore
backward-forward provides a more computational efficient optimization
framework for distributed MTL. In addition, we note that the backward-forward
iteration is a non-expansive operator if $\eta\in(0,2/L)$ because both the
forward and backward steps are non-expansive.


When applying the coordinate update scheme in~\cite{Arock}, all task nodes have
access to the shared memory, and they do not communicate with each other. Further
the communicate between each task node and the central server is only the vector
$v_t$ , which is typically much smaller than the data $\mathcal D_t$ stored
locally at each task node. In the proposed AMTL scheme, the task nodes do not
share memory but are exclusively connected and communicate with the central
node. The computation of the backward step is located in the central node,
which performs the proximal mapping after one gradient update is received from
a task node (the proximal mapping can be also applied after several gradient updates depending on the speed of gradient update). In this case, we further save the communication cost between each
task node and the central node, because each task node only need the task block
corresponding to the task node. 

\begin{figure}[t!]
\center
\includegraphics[width=0.5\textwidth]{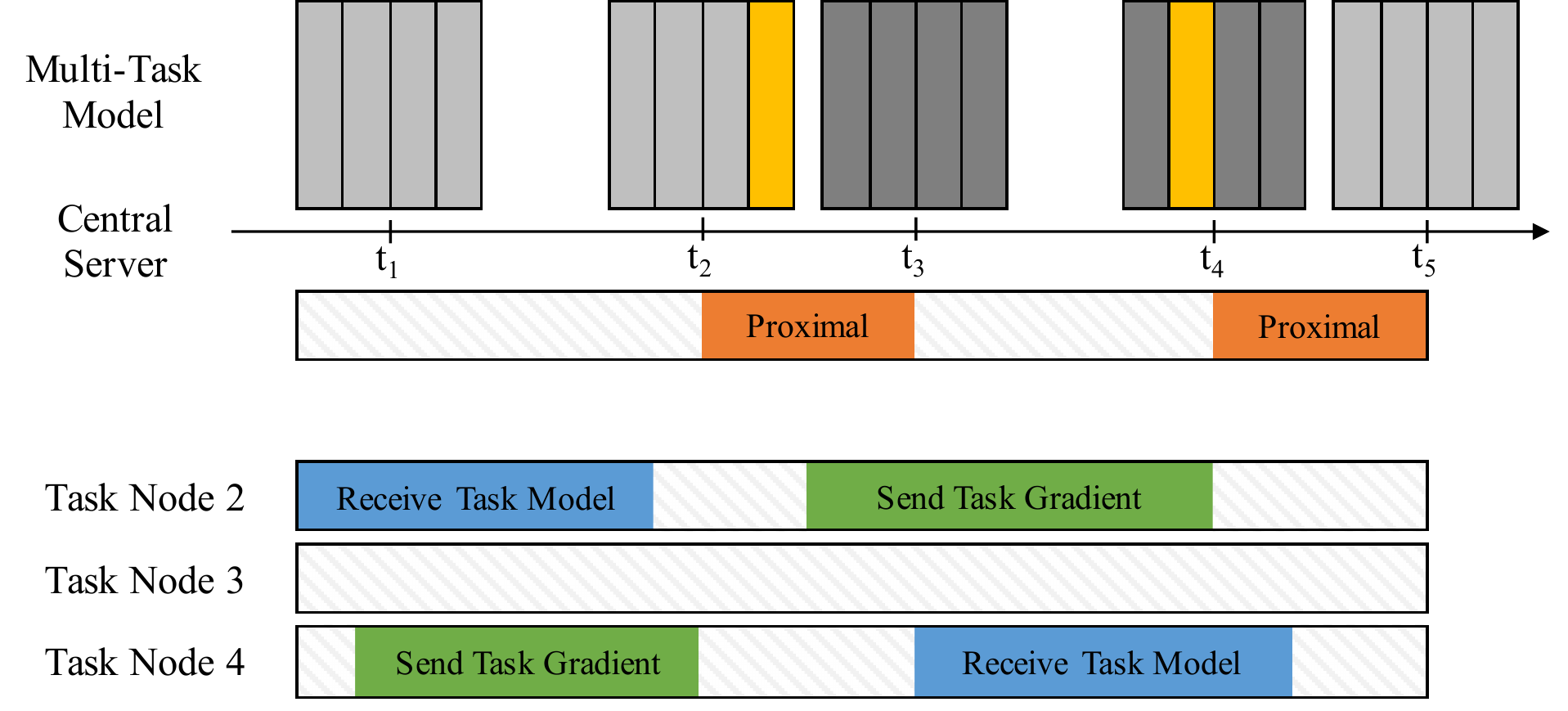}
\caption{Illustration of asynchronous updates in AMTL. 
The asynchronous update scheme has an inconsistency when it
comes to read model vectors from the central server. Such
inconsistencies caused by the backward step of the AMTL is taken into
account in the convergence analysis.}
\vspace{-0.1in}
\label{fig:async_update_fig}
\end{figure}

To illustrate the asynchronous update mechanism in AMTL, we show an example in
Fig.~\ref{fig:async_update_fig}. The figure shows order of the backward and
the forward steps performed by the central node and the task nodes. At time
$t_1$, the task node 2 receives the model corresponding to the task 2 which
was previously updated by the proximal mapping step in the central node. As
soon as task node~2 receives its model, the forward (gradient) step is
launched. After the task gradient descent update is done, the model of the task~2
is sent back to the central node. When the central node receives the updated
model from the task node~2, it starts applying the proximal step on the whole
multi-task model matrix. As we can see from the figure, while task node~2 was
performing its gradient step, task node~4 had already sent its updated model
to the central node and triggered a proximal mapping step during time steps $t_2$ and $t_3$. Therefore, the model matrix was updated upon the request of
the task node~4 during the gradient computation of task node 2. Thus we know
that the model received by the task node~2 at time $t_1$ is not the same copy
as in the central server any more. 
When the updated model from the task node 2 is received by the central node,
proximal mapping computations are done by using the model received from task
node~2 and the updated models at the end of the proximal step triggered by the
task node 4. Similarly, if we think of the model received by task node~4 at
time $t_3$, we can say that it will not be the same model as in the central
server when task node 4 is ready to send its model to the central node
because of the proximal step triggered by the task node~2 during time steps $t_4$ and $t_5$. This is because in AMTL, there is no memory lock during reads.
As we can see, the asynchronous update scheme has inconsistency when it
comes to read model vectors from the central server. We note that such
inconsistency caused by the backward step is already taken into
account in the convergence analysis.


We summarize the proposed AMTL algorithm in
Algorithm~\ref{alg:amtl}. The AMTL framework 
enjoys the following convergence property:

\begin{theorem}
Let $(V^k)_{k\geq0}$ be the sequence generated by the proposed AMTL with
$\eta_k \in [\eta_{\min}, \frac{c}{2\tau/\sqrt{T}+1} ]$ for any
$\eta_{\min}>0$ and $0<c<1$, where $\tau$ is the maximum delay. Then $(V^k)_{k\geq 0}$ converges to an
$V^*$-valued random variable almost surely. If the MTL problem in
Eq.~\ref{eqt:mtl} has a unique solution, then the sequence converges to the
unique solution.
\end{theorem}

According to our assumptions, all the task nodes are independent Poisson
processes and each task node has the same activation rate. The probability
that each task node is activated before other task nodes is
$1/T$~\cite{larson1981urban}, and therefore we can assume that each coordinate
is selected with the same probability. The results in~\cite{Arock} can be
applied to directly obtain the convergence. We note that some MTL algorithms based on
sparsity-inducing norms may not have a unique solution, as commonly seen in
many sparse learning formulations, but in this case we typically can add an
$\ell_2$ term to ensure the strict convexity and obtain linear convergence as shown in~\cite{Arock}. An example of such technique is
the elastic net variant from the original Lasso
problem~\cite{zou2005regularization}.

\subsection{Dynamic step size controlling in AMTL}
\label{sect:method:step_size}

As discussed in Remark~\ref{rmk:act_rate}, the AMTL is based on the same
activation rate in Assumption~\ref{asp:act_rate}. However because of the
topology of the network in real-world settings, the same activation rate is
almost impossible. In this section, we propose a dynamic step size for AMTL to
overcome this challenge. We note that in order to ensure convergence, the step
sizes used in asynchronous optimization algorithms are typically much smaller
than those used in the synchronous optimization algorithms, which
limits the algorithmic efficiency of the solvers. The dynamic step was recently
used in a specific setting via asynchronous optimization to achieve
better overall performance~\cite{cheung2014amortized}. 

Our dynamic step size is motivated by this design, and revises the update of AMTL in 
Eq.~\ref{eq:update} by augmenting a time related multiplier:
\begin{equation}
	\begin{split}
	v_t^{k+1} &= v_t^k + c_{(t, k)}\eta_k \left(\left(\mbox{Prox}_{\tau \lambda g
}\left(\hat{{v}}^{k}\right)\right)_t \vphantom{\sum_{t=1}^{T}} \right. \\ 
	                & \left. -  \eta   \nabla \ell_t
	\left(\left(\mbox{Prox}_{\tau \lambda g}
	(\hat{{v}}^{k})\right)_t\right)-v_t^k\right)
	\end{split}
	\label{eq:update-dynamic}
\end{equation}%
where the multiplier is given by:
\begin{equation}
c_{(t, k)} = \log \left(\max \left(\bar \nu_{t, k},10\right)\right)
\label{eq:transformation}
\end{equation}
and $\bar \nu_{t, k} = \tfrac{1}{k+1}\sum_{i=z-k}^{z} \nu_t^{(i)} $ is the
average of the last $k+1$ delays in task node $t$, $z$ is the current time point,
and $\nu_t^{(i)}$ is the delay at time $i$ for task $t$. As such, the actual
step size given by $c_{(t, k)} \eta_k$ will be scaled by the history of
communication delay between the task nodes and the central server. The longer
the delay, the larger is the step size to compensate for the loss from
the activation rate. We note that though in our experiments we show the
effectiveness of the propose scheme, instead of using
Eq.~(\ref{eq:transformation}), different kinds of function could also be used.
Currently, these are no theoretical results on how a dynamic step could improve
the general problem. Further what types of dynamics could better serve
the purpose remains an open problem in the optimization research.


\section{Numerical Experiments}
\label{sect:exp}

The proposed asynchronous distributed MTL framework is implemented
in C++. In our experiments, we simulate the distributed environment using
the shared memory architecture in~\cite{Arock} with network delays introduced
to the work nodes. In this section, we first elaborate how AMTL performs and then present experiments on synthetic and
real-world datasets to compare the training times of the proposed AMTL and synchronous distributed multi-task learning (SMTL). We also investigate empirical convergence behaviors 
of AMTL and traditional SMTL by comparing them on synthetic datasets. Finally, we study the effect of dynamic
step size on synthetic datasets with various numbers of tasks.
Experiments were conducted with an Intel Core i5-5200U CPU 2.20GHz x 4 laptop.
Note that the performance is limited by the hardware specifications such as the
number of cores.

\subsection{Experimental setting} 

In order to simulate the distributed environment using a shared memory system,
we use threads to simulate the task nodes, and the number of threads is equal to
the number of tasks. As such, each thread is responsible for learning the model parameters of one task and communicating with the central node to achieve knowledge transfer, where the central node is simulated by the shared memory. 

Though the framework can be used to solve many regularized MTL formulations,
in this paper we focus on one specific MTL formulation--the low-rank MTL for shared
subspace learning. In the formulation, we assume that all tasks learn a regression
model with the least squares loss $\sum_{t=1}^T \left\|x_{t}w_{t} -
y_{t}\right\|_{2}^{2}$. Recall that $x_{t},~n_{t},~x_{t,i}$, and $y_{t,i}$
denote the data matrix, sample size, $i$-th data sample of the task $y$, and the $i$-th label of
the task $t$, respectively. The nuclear norm is used as the regularization
function that couples the tasks by learning a shared low-dimensional subspace,
which serves as the basis for knowledge transfer among tasks. The formulation 
is given by:
\begin{align}
\min_{W} \left\{  \sum\nolimits_{t=1}^T \left\|x_{t}w_{t} - y_{t}\right\|_{2}^{2} + \lambda \|W\|_* \right\}.
\end{align}

As it was discussed in the previous sections, we used the backward-forward
splitting scheme for AMTL, since the nuclear norm is not separable. The
proximal mapping for nuclear norm regularization is given
by~\cite{ji2009accelerated,cai2010singular}:
\begin{equation}
\begin{split}
\mbox{Prox}_{\eta \lambda g} \left(\hat{{V}}^{k}\right) &= \sum_{i=1}^{\mbox{min} \left\{d,T\right\}} \mbox{max} \left(0, \sigma_{i} - \eta \lambda\right)u_{i}v_{i}^{\top} \\
&= U \left(\Sigma - \eta \lambda I\right)_{+} V^{\top}
\end{split}
\label{eq:soft_thresholding}
\end{equation}
where $\{u_i\}$ and $\{v_i\}$ are columns of $U$ and $V$, respectively, $\hat{{V}}^{k} = U \Sigma V^{\top}$ is the singular value decomposition
(SVD) of $\hat{{V}}^{k}$ and $\left(x\right)_{+} =
\mbox{max}\left(0,x\right)$. In every stage of the AMTL framework,
copies of all the model vectors are stored in the shared memory. Therefore,
when the central node is performing the backward step, it retrieves the current
versions of the models from the shared memory. Since the proposed framework is
an asynchronous distributed system, copies of the models may be changed by
task nodes while the central node is performing the proximal mapping. Whenever a task node completes its computation, it sends its model to the central
node for proximal mapping without waiting for other task nodes to finish their
forward steps.

As it is seen in Eq. \eqref{eq:soft_thresholding}, every backward step
requires a singular value decomposition (SVD) of the model matrix. When the
number of tasks $T$ and the dimensionality $d$ are high, SVD is a
computationally expensive operation. Instead of computing the full SVD at
every step, online SVD can also be used~\cite{brand2003fast}. Online SVD updates $U,~V$, and $\Sigma$
matrices by using the previous values. Therefore, SVD is performed once
at the beginning and those left, right, and singular value matrices are used to
compute the SVD of the updated model matrix. Whenever a column of the model
matrix is updated by a task node, central node computes the proximal mapping.
Therefore, instead of performing the full SVD every time, we can update
the SVD of the model matrix according to the changed column. When we need to
deal with a huge number of tasks and high dimensionality, online SVD can be used
to mitigate computational complexity.

 
\subsection{Comparison between AMTL and SMTL}


\subsubsection{Public and Synthetic Datasets}

\begin{figure*}[t!]
\centering
\subfloat[Varying number of tasks]{
\includegraphics[width=0.32\textwidth]{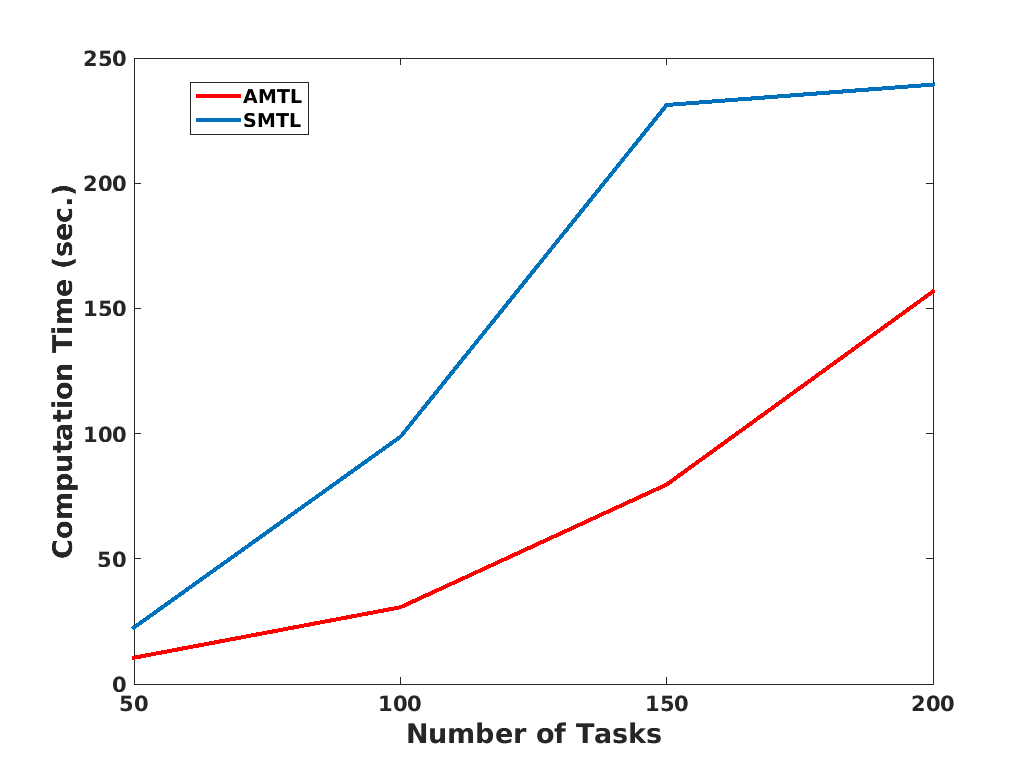}
\label{tasks}}
\subfloat[Varying sample sizes in each task]{
\includegraphics[width=0.32\textwidth]{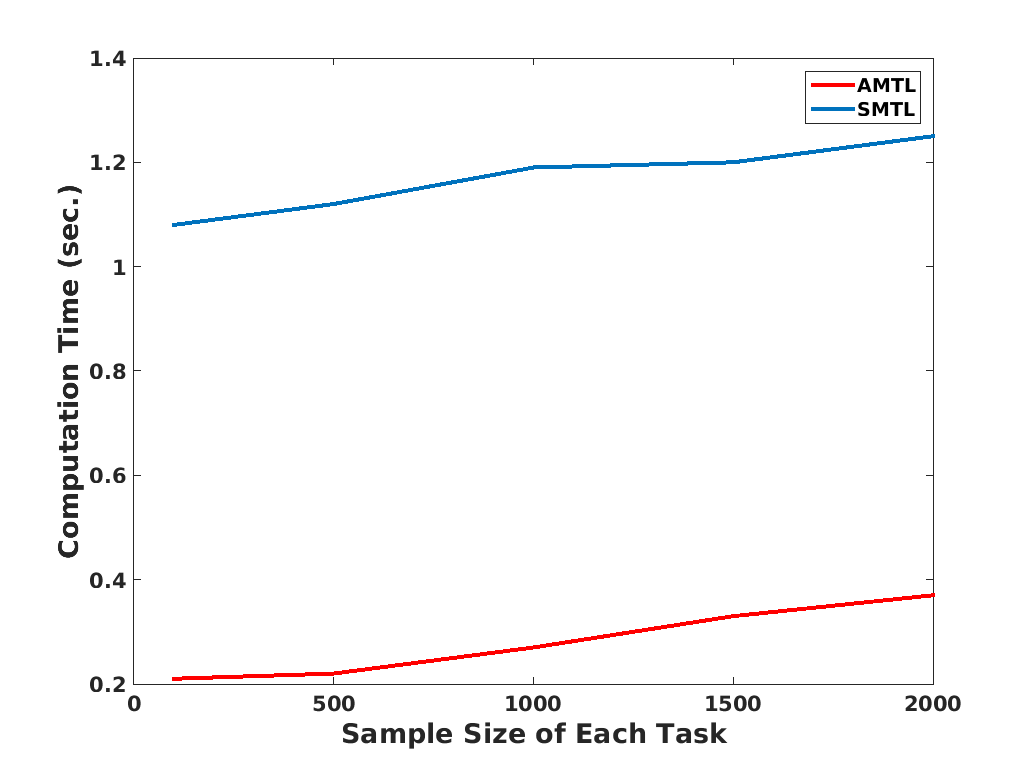}
\label{sizes}}
\subfloat[Varying dimensionalities of the model]{
\includegraphics[width=0.32\textwidth]{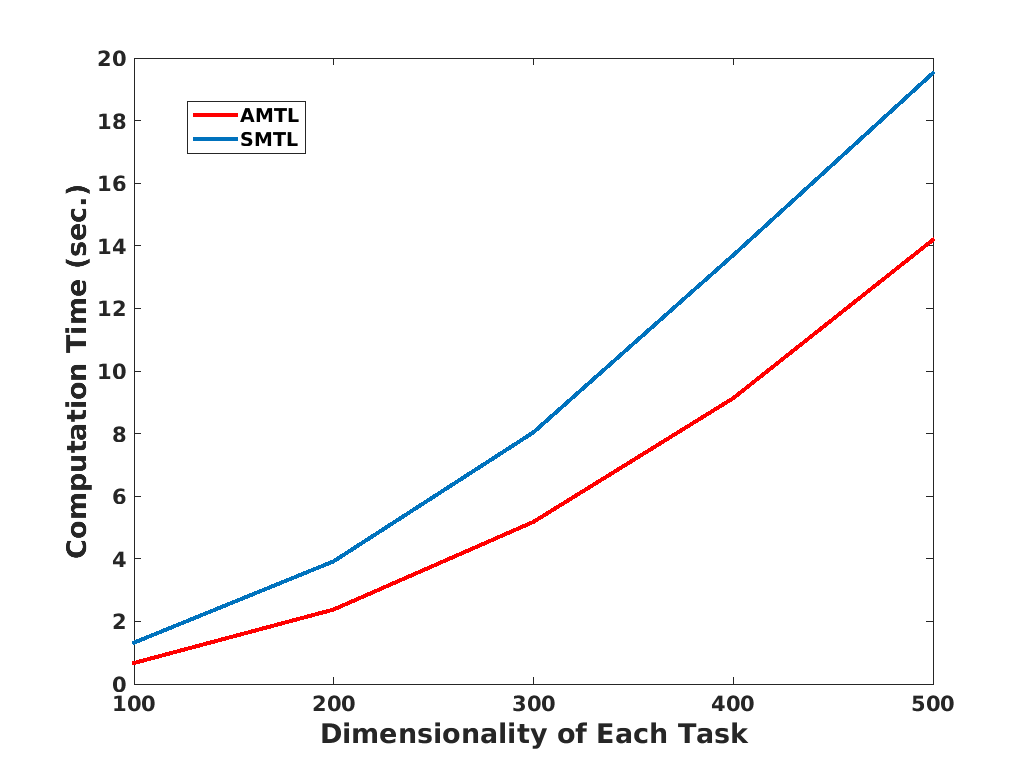}
\label{dim}}
\caption{Computation times of AMTL and SMTL for a) varying number of tasks for $50$ dimensional $100$ samples in each task, b) varying number of task sizes with $50$ dimensions, and $5$ tasks, c) varying dimensionality of $5$ tasks with $100$ samples in each. SMTL requires more computational time than AMTL for a fixed number of
	iterations.}
\label{fig:async_vs_sync}
\end{figure*}

In this section, the difference in computation times of AMTL and SMTL is
investigated with varying number of tasks, dimensionality, and sample sizes since both AMTL and SMTL have nearly identical progress per iteration (every task node updates one forward step for each iteration).
Synthetic datasets were randomly generated. In
Fig.~\ref{fig:async_vs_sync}, the computation time for a varying number of
tasks, sample sizes, and dimensionalities is shown. In Fig.~\ref{tasks}, the dimensionality of the dataset was chosen as $50$, and the
sample size of each task was chosen as $100$. As observed from
Fig.~\ref{tasks}, computation time increases with increasing
number of tasks because the total computational complexity increases. However, the increase is more drastic for
SMTL. Since each task node has to wait for other task nodes to finish their
computations in SMTL, increasing the number of tasks causes the algorithm to
take much longer than AMTL. For Fig.~\ref{tasks}, computation time still increases after $100$ tasks for AMTL because the number of backward steps increases as the number of tasks increases. An important point we should note is that the number of cores we ran our experiments was less than $100$. There is a dependecy on hardware.

In the next experiment, random datasets were generated with different sample
sizes, and the effect of varying sample sizes on the computation times of AMTL
and SMTL is shown in Fig.~\ref{sizes}. The number of
tasks were chosen as $5$, and the dimensionality was $50$. Increasing the
sample size did not cause abrupt changes in computation times for both
asynchronous and synchronous settings. That is because computing the gradient has a similar cost to the proximal mapping for small sample sizes and the cost for computing the gradient increases as the sample size increases while the cost for the proximal mapping keeps unchanged.
However, AMTL is still faster than SMTL
even for a small number of tasks.

In Fig.~\ref{dim}, the
computational times of AMTL and SMTL for different dimensionalities is shown. As 
expected, the time required for both schemes increases with higher
dimensionalities. On the other hand, we can observe from
Fig.~\ref{dim} that the gap between AMTL and
SMTL also increases. In SMTL, task nodes have to wait longer for the updates,
since the calculations of the backward and the forward steps are prolonged
because of higher $d$.

In Table~\ref{tab:summary}, computation times of AMTL and SMTL with different
network characteristics for synthetic datasets with a varying number of tasks
are summarized. Synthetic datasets were randomly generated with $100$ samples
for each task, and the dimensionality was set to $50$. The results are shown for
datasets with $5, 10$, and $15$ tasks. Similar to the previous
experimental settings, a regression problem with the squared loss and nuclear norm
regularization was taken into account. As it was shown, AMTL outperforms SMTL
at every dimensionality, sample size, and number of tasks considered here. AMTL also
performs better than SMTL under different communication delay patterns. In Table~\ref{tab:summary}, AMTL-5, AMTL-10, and AMTL-30 represent the AMTL framework
where the offset values of the delay were chosen as 5, 10, and 30 seconds.
Simulating the different network settings for our experiments, an offset
parameter was taken as an input from the user. This parameter represents an
average delay related to the infrastructure of the network. Then the amount of
delay was computed as the sum of the offset and a random value in each task
node. AMTL is shown to be more time efficient than SMTL under the same network
settings.

\begin{table}[!t]
\small
\caption{Computation times (sec.) of AMTL and SMTL with different network
characteristics. The offset value of the delay for AMTL-5, 10, 30 was chosen as 5, 10, 30 seconds. Same network settings were used to
compare the performance of AMTL and SMTL. AMTL performed better than SMTL for
all the network settings and numbers of tasks considered here.}
\label{tab:summary}
\centering
\begin{tabular}{| c | c | c | c |}
\hline
 Network & 5 Tasks & 10 Tasks & 15 Tasks \\  
\hline
AMTL-5 & 156.21 & 172.59 & 173.38 \\
\hline
AMTL-10 & 297.34 & 308.55 & 313.54 \\
\hline
AMTL-30 & 902.22 & 910.39 &  880.63\\
\hline
SMTL-5 & 239.34 & 248.23 & 256.94 \\
\hline
SMTL-10 & 452.84 & 470.79 & 494.13\\
\hline
SMTL-30 & 1238.16 & 1367.38 & 1454.57\\
\hline
\end{tabular}
\vspace{-0.07in}
\end{table} 

The performance of AMTL and SMTL is also shown for three public datasets.
The number of tasks, sample sizes, and the dimensionality of the data sets are
given in Table~\ref{tab:table_datasets}. School and MNIST are commonly used
public datasets. School dataset has exam records of 139 schools in 1985, 1986, 
and 1987 provided by the London Education Authority (ILEA) \cite{school}. MNIST is a popular handwritten digits dataset with $60,000$ 
training samples and $10,000$ test samples \cite{MNIST}. MNIST is prepared as $5$ binary
classification tasks such as $0$ vs $9$, $1$ vs $8$, $2$ vs $7$, $3$ vs $6$,
and $4$ vs $5$ . Another public dataset used in experiments was Multi-Task
Facial Landmark (MTFL) dataset~\cite{MTFL}. In this dataset, there are
$12,995$ face images with different genders, head poses, and characteristics.
The features are five facial landmarks, attribute of gender, smiling/not
smiling, wearing glasses/not wearing glasses, and head pose. We designed four
binary classification tasks such as male/female, smiling/not smiling,
wearing/not wearing glasses, and right/left head pose. The logistic loss 
was used for binary classification tasks, and the squared loss was used for the
regression task. Training times of AMTL and SMTL are given in Table
\ref{tab:table_computation_time}. When the number of tasks is high, the gap
between training times of AMTL and SMTL is wider. The training time of AMTL is
always less than the training time of SMTL for all of the real-world datasets
for a fixed amount of delay in this experiment.

\begin{table}[!t]
\small
\caption{Benchmark public datasets used in this paper.}
\label{tab:table_datasets}
\centering
\begin{tabular}{| c | c | c | c |}
\hline
 Data set & Number of tasks & Sample sizes & Dimensionality \\  
\hline
\hline
School & 139 & 22-251 & 28 \\
\hline
MNIST & 5 & 13137-14702 & 100 \\
\hline
MTFL & 4 & 2224-10000 & 10 \\
\hline 
\end{tabular}
\end{table}

\begin{table}[!t]
\small
\caption{Training time (sec.) comparison of AMTL and SMTL for public datasets.
Training time of AMTL is less than the training time of SMTL for real-world
datasets with different network settings.}
\label{tab:table_computation_time}
\centering
\begin{tabular}{| c | c | c | c |}
\hline
 Network & School  & MNIST & MTFL  \\  
\hline
\hline
AMTL-1 & 194.22 & 54.96 & 50.40 \\
\hline
AMTL-2 & 231.58 & 83.17 & 77.44 \\
\hline
AMTL-3 & 460.15 & 115.46 & 103.45 \\
\hline
SMTL-1 & 299.79 & 57.94 & 50.59 \\
\hline
SMTL-2 & 298.42 & 114.85 & 92.84 \\
\hline
SMTL-3 & 593.36 & 161.67 & 146.87 \\
\hline
\end{tabular}
\vspace{-0.06in}
\end{table}

\begin{figure}[t!]
	\vspace{-0.3in}
	\centering
		\includegraphics[width=0.4\textwidth]{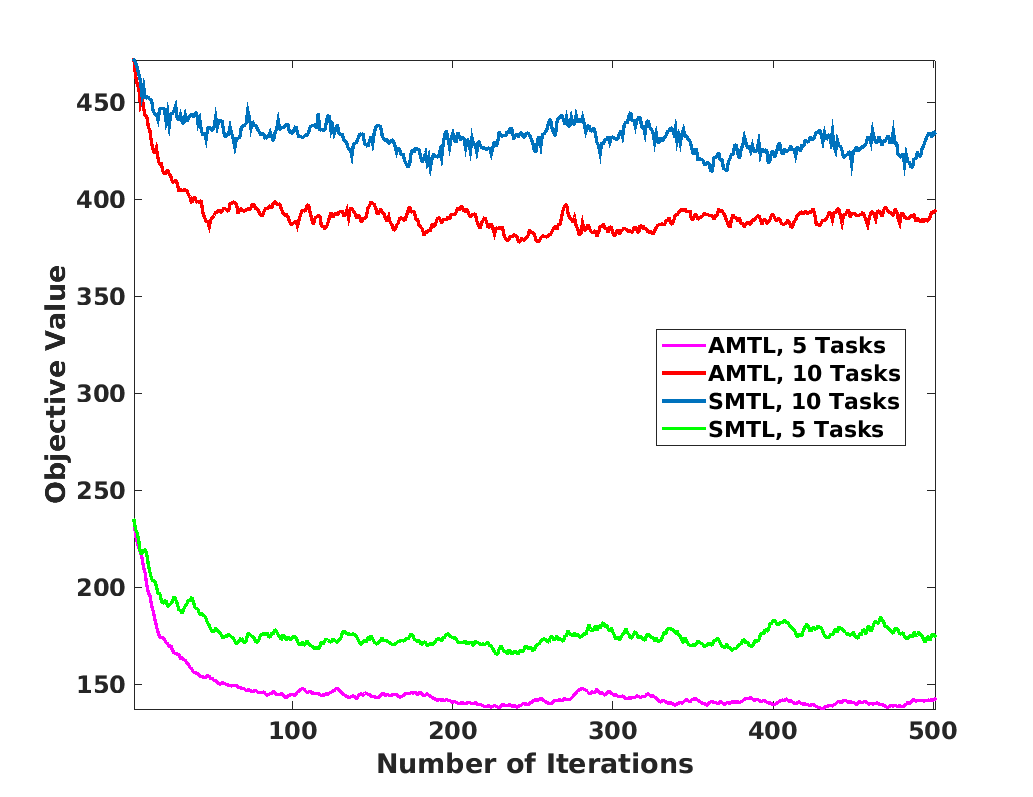}
	\caption{Convergence of AMTL and STML under the same network configurations. Experiment was conducted for randomly generated synthetic datasets with $5$ and $10$ tasks. AMTL is not only more time efficient than SMTL, it also tends to converge faster than STML.}
	\label{fig:convergence}
\end{figure}

Experiments show that AMTL is more time efficient than SMTL, especially,
when there are delays due to the network communication. In this situation, asynchronous
algorithm becomes a necessity, because network communication increases the
training time drastically. Since each task node in AMTL performs the backward-forward splitting steps and the variable updates without waiting for any other
node in the network, it outperforms SMTL for both synthetic and real-world
datasets with a various number of tasks and network settings. Moreover, AMTL
does not need to carry the raw data samples to a central node. Therefore, it is
very suitable for private datasets located at different data centers compared
to many distributed frameworks in the literature. In addition to time efficiency, convergence of AMTL and SMTL under same network configurations are compared. In Fig.~\ref{fig:convergence}, convergence curves of AMTL and SMTL are given for a fixed number of iterations and synthetic data with $5$ and $10$ tasks. As seen in the figure, AMTL tends to converge faster than SMTL in terms of the number of iteratios as well.

\vspace{-0.1in}

\subsection{Dynamic step size}

In this section, we present the experimental result of the proposed dynamic
step size. A dynamic step size is proposed by utilizing the delays in the
network. We simulate the delays caused by the network communication by keeping
the task nodes idle for while after it completes the forward step. The
dynamic step size was computed by Eq.~(\ref{eq:transformation}). In this
experiment, the average delay of the last 5 iterations was used to modify the step
size. The effect of the dynamic step size was shown for randomly generated
synthetic datasets with $100$ samples in each task with the dimensionality was set to $50$. The objective values of each dataset with different numbers of tasks and
with different offset values were examined. These objective values were
calculated at the end of the total number of iterations and the final updated
versions of the model vectors were used. Because of the delays, some of the
task nodes have to wait much longer than other nodes, and the convergence slows
down for these nodes. If we increase the step size of the nodes which had to
wait for a long time in previous iterations, we can boost the convergence. The
experimental results are summarized in Tables
\ref{tab:dynamic1},~\ref{tab:dynamic2}, and
\ref{tab:dynamic3}. According to our observation, if we use dynamic step size,
the objective value decreases compared with the objective value of the AMTL
with constant step size. The convergence criteria was chosen as a fixed number
of iterations such as~$10$. As we can see from the tables, the dynamic step
size helps to speed up the convergence. Another observation is that objective
values tend to decrease with the increasing amount of delay, when the dynamic
step size is used. Although no theoretical results are available to quantify the dynamic step size in existing literature, the empirical results we obtained indicate that the dynamic step size is a very promising strategy and is especially effective when there are significant delays in the network.

\begin{table}[!t]
\vspace{-0.1in}
\caption{Objective values of the synthetic dataset with 5 tasks under different network settings.}
\small
\label{tab:dynamic1}
\centering
\begin{tabular}{| c | c | c |}
\hline
 Network & Without dynamic step size  & Dynamic step size  \\  
\hline
\hline
AMTL-5 &  163.62 & 144.83\\
\hline
AMTL-10 &  163.59 & 144.77\\
\hline
AMTL-15 &  163.56 & 143.82\\
\hline
AMTL-20 &  168.63 & 143.50\\
\hline
\end{tabular}
\end{table}

\begin{table}[!t]
\caption{Objective values of the synthetic dataset with 10 tasks. Objective values are shown for different network settings. Dynamic step size yields lower objective values at the end of the last iteration than fixed step size.}
\small
\label{tab:dynamic2}
\centering
\begin{tabular}{| c | c | c |}
\hline
 Network & Without dynamic step size  & Dynamic step size  \\  
\hline
\hline
AMTL-5 & 366.27  & 334.24\\
\hline
AMTL-10 & 367.63 & 333.71 \\
\hline
AMTL-15 & 366.26 & 333.12  \\
\hline
AMTL-20 & 366.35 & 331.13 \\
\hline
\end{tabular}
\end{table}

\begin{table}[!t]
\caption{Objective values of the synthetic dataset with 15 tasks. The difference between objective values of AMTL with and without dynamic step size is more visible when the amount of delay increases.}
\small
\label{tab:dynamic3}
\centering
\begin{tabular}{| c | c | c |}
\hline
 Network & Without dynamic step size  & Dynamic step size  \\  
\hline
\hline
AMTL-5 & 559.07 & 508.65  \\
\hline
AMTL-10 & 561.68 & 505.64 \\
\hline
AMTL-15 & 561.87 & 500.05\\
\hline
AMTL-20 & 561.21 & 499.97\\
\hline
\end{tabular}
\vspace{-0.06in}
\end{table}

\section{Conclusion}
\label{sect:conclusion}
In conclusion, a distributed regularized multi-task learning approach is
presented in this paper. An asynchronous distributed coordinate update method
is adopted to perform full updates on model vectors. Compared to other
distributed MTL approaches, AMTL is more time efficient because task nodes do not need to wait for other nodes to perform the gradient updates. 
A dynamic step size to boost the convergence performance is investigated by
scaling the step size according to the delays in the communication. Training
times are compared for several synthetic, and public datasets and the results
showed that the proposed AMTL is faster than traditional synchronous MTL. 
We also study the convergence behavior of AMTL and SMTL by comparing the precision of the two approaches. 
We note that current AMTL implementation is based on the ARock~\cite{Arock} framework, which largely limits our capability of conducting experiments for different network structures. As our future work, we will develop
a standalone AMTL implementation\footnote{Available at \url{https://github.com/illidanlab/AMTL}} that allows us to validate AMTL in real-world network settings. Stochastic gradient approach will also be incorporated into the current distributed
AMTL setting.


\section*{Acknowledgment}
This research is supported in part by the National Science Foundation (NSF) under grant
numbers IIS-1565596, IIS-1615597, and DMS-1621798 and the Office of Naval Research (ONR) under grant number
N00014-14-1-0631.



\bibliographystyle{IEEEtran}
\bibliography{IEEEabrv,icdm_bib}

%

\end{document}